\documentclass[sigconf]{aamas}

\usepackage{amsmath}
\usepackage{graphicx}
\usepackage{subcaption}
\usepackage{xcolor} 
\usepackage{hyperref} 
\usepackage{color}
\usepackage{xspace}
\usepackage{array}
\usepackage{seqsplit}
\usepackage{epsfig}
\usepackage{multirow}
\usepackage{url}
\usepackage{booktabs}
\usepackage{balance}
\usepackage{pifont}
\usepackage{rotating}
\usepackage{tikz}
\usetikzlibrary{positioning, arrows.meta}
\usepackage{placeins}
\usepackage{enumitem}
\usepackage{arydshln}
\usepackage{tabularx}
\usepackage{enumitem}
\usepackage{titlesec}

\newcommand{\myparagraph}[1]{\par\noindent\textbf{#1}. }
\newcommand{\firstmention}[1]{{\bf #1}}

\usepackage{amsmath,amsfonts,bm}

\def\eqref#1{equation~\ref{#1}}

\def\1{\bm{1}}

\DeclareMathAlphabet{\mathsfit}{\encodingdefault}{\sfdefault}{m}{sl}
\SetMathAlphabet{\mathsfit}{bold}{\encodingdefault}{\sfdefault}{bx}{n}

\titlespacing*{\section}
{0pt}{1.0ex plus 0.5ex minus 0.5ex}{0.8ex}
\titlespacing*{\subsection}
{0pt}{0.8ex plus 0.3ex minus 0.3ex}{0.6ex}
\titlespacing*{\subsubsection}
{0pt}{0.6ex}{0.4ex}
\setlist{nosep}
\setcopyright{ifaamas}
\acmConference[AAMAS '26]{Proc.\@ of the 25th International Conference
on Autonomous Agents and Multiagent Systems (AAMAS 2026)}{May 25 -- 29, 2026}
{Paphos, Cyprus}{C.~Amato, L.~Dennis, V.~Mascardi, J.~Thangarajah (eds.)}
\copyrightyear{2026}
\acmYear{2026}
\acmDOI{}
\acmPrice{}
\acmISBN{}

\acmSubmissionID{1241}

\title{LLM Performance Predictors: Learning When to Escalate in Hybrid Human-AI Moderation Systems}
\author{Or Bachar}
\affiliation{%
  \institution{Zefr}
  \city{Los Angeles}
  \country{United States}
}
\additionalaffiliation{%
  \institution{Reichman University}
  \city{Herzliya}
  \country{Israel}
}
\email{or.bachar@zefr.com}
\email{or.bachar@post.runi.ac.il}

\author{Or Levi}
\affiliation{
  \institution{Zefr}
  \city{Los Angeles}
  \country{United States}
  }
\email{or.levi@zefr.com}

\author{Sardhendu Mishra}
\affiliation{
  \institution{Zefr}
  \city{Los Angeles}
  \country{United States}
  }
\email{sardhendu.mishra@zefr.com}

\author{Adi Levi}
\affiliation{
  \institution{Zefr}
  \city{Los Angeles}
  \country{United States}
  }
\email{adi.levi@zefr.com}

\author{Manpreet Singh Minhas}
\affiliation{
  \institution{Zefr}
  \city{Los Angeles}
  \country{United States}
  }
\email{manpreet.minhas@zefr.com}

\author{Justin Miller}
\affiliation{
  \institution{Zefr}
  \city{Los Angeles}
  \country{United States}
  }
\email{justin.miller@zefr.com}

\author{Omer Ben-Porat}
\affiliation{
  \institution{Technion}
  \city{Haifa}
  \country{Israel}
  }
\email{omerbp@technion.ac.il}

\author{Eilon Sheetrit}
\affiliation{
  \institution{Reichman University}
  \city{Herzliya}
  \country{Israel}
  }
\email{eilon.sheetrit@post.runi.ac.il}

\author{Jonathan Morra}
\affiliation{
  \institution{Zefr}
  \city{Los Angeles}
  \country{United States}
  }
\email{jon.morra@zefr.com}

\begin{abstract}
As LLMs are increasingly integrated into human-in-the-loop content moderation systems, a central challenge is deciding when their outputs can be trusted versus when escalation for human review is preferable. We propose a novel framework for supervised LLM uncertainty quantification, learning a dedicated meta-model based on LLM Performance Predictors (LPPs) derived from LLM outputs: log-probabilities, entropy, and novel uncertainty attribution indicators.
We demonstrate that our method enables cost-aware selective classification in real-world human-AI workflows: escalating high-risk cases while automating the rest. Experiments across state-of-the-art LLMs, including both off-the-shelf (Gemini, GPT) and open-source (Llama, Qwen), on multimodal and multilingual moderation tasks, show significant improvements over existing uncertainty estimators in accuracy-cost trade-offs. Beyond uncertainty estimation, the LPPs enhance explainability by providing new insights into failure conditions (e.g., ambiguous content vs. under-specified policy). This work establishes a principled framework for uncertainty-aware, scalable, and responsible human-AI moderation workflows.
\end{abstract}
\newcommand{\BibTeX}{\rm B\kern-.05em{\sc i\kern-.025em b}\kern-.08em\TeX}

\makeatletter
\providecommand{\@acmBadgeL@image}{}
\providecommand{\@acmBadgeR@image}{}
\makeatother

\begin{document}


\pagestyle{fancy}
\fancyhead{}


\maketitle 


\section{Introduction}

Human–AI collaboration is at the heart of modern content moderation, which serves as a cornerstone of online trust and safety. Effective moderation is essential for protecting individuals from harmful or misleading content, and the pursuit of scalable and responsible approaches contributes to a safer online environment. To meet this need, moderation systems must balance the nuanced accuracy of human experts with the scalability and cost-efficiency of AI. The rapid growth of user-generated content has made reliance solely on manual review infeasible, motivating the growing deployment of Large Language Models (LLMs) into moderation workflows. When LLMs are used in human-in-the-loop workflows, their utility depends not only on their accuracy, but also on their ability to signal when they are likely to make an incorrect decision.  The effectiveness of such pipelines depends on a single decision that repeats billions of times per day: should the system trust the LLM's judgment or escalate to a human reviewer?

Accordingly, we focus on two research questions:\\
\noindent \firstmention{RQ1:} Can we accurately predict when LLMs are likely to be incorrect in moderation tasks? \\
\noindent \firstmention{RQ2:} Can selective escalation based on these predictions reduce the overall cost in human-AI moderation workflows?

This trust-or-escalate decision is, at its core, an uncertainty estimation problem. 
Recent work~\cite{shorinwa2025surveyuncertaintyquantificationlarge,  farr+al:25, kapoor2024large, cohen2024i, efficient-and-effective, liu+al:24} has explored a variety of signals for estimating LLM uncertainty. Approaches range from token-level entropy and log-probability thresholds~\cite{jiang+al:21, huly+al:25} to ensemble-based sampling~\cite{farr+al:25} and explicit self-expressed confidence~\cite{tian+al:23, kadavath+al:22, lin+al:22, xiong+al:23}.

In the realm of content moderation, current LLM uncertainty metrics fail to capture the economic value of moderation failure. To address this, we formulate the decision as a cost minimization problem, balancing between the cost of misclassification and the cost of human review. Moreover, previous work~\cite{shorinwa2025surveyuncertaintyquantificationlarge,jiang+al:21} showed that no single universal uncertainty metric has been proven to be reliable across tasks and models. These findings motivate the pursuit of a robust approach that can perform reliably at scale in high-stakes domains such as content moderation. 

To this end, we introduce a novel framework that builds on this foundation, enabling a unified approach to selective escalation. Unlike prior supervised uncertainty estimation approaches that typically rely on single or homogeneous signals~\cite{liu+al:24, jiang+al:21, hendrycks2017a}, or learning-to-defer methods that optimize deferral without uncertainty attribution~\cite{geifman2017selectiveclassificationdeepneural, mozannar2021consistentestimatorslearningdefer}, we fuse diverse gray- and black-box output signals and incorporate moderation-specific abstention with attribution, producing more comprehensive and explainable LLM Performance Predictors (LPPs). The approach is inspired by Query Performance Prediction (QPP)~\cite{carmel+Elad:10,hauff+al:08,shtok+al:16,shtok+al:12,tao+al:14,zhou+croft:07} in Information Retrieval, which studies how systems can learn to anticipate the success of individual queries. QPPs are commonly divided into pre-retrieval predictors and post-retrieval predictors. In this work, we focus on post-generation predictors, which are more scarce in the literature~\cite{huly+al:25}, due to strong empirical predictive power, as they leverage richer evidence from the actual model outputs and broad applicability for both open and closed models.

Uncertainty estimators for LLMs are often categorized by the level of access to model internals: white-box approaches leverage hidden states and gradients, gray-box approaches use output-side signals such as log-probabilities, and black-box approaches rely only on final predictions or agreement across multiple generations. In this work, we focus on gray- and black-box features, as they strike a practical balance between richness of signal, scale, and applicability across both off-the-shelf and open-source LLMs.

Beyond estimating the level of uncertainty, moderation teams also need to know the reason for the uncertainty. We therefore introduce novel moderation-oriented uncertainty attribution indicators that distinguish evidence deficits (e.g. missing transcript context, visually ambiguous frames, cross-lingual inconsistencies) from policy gaps (e.g., edge cases not covered by moderation guidelines, conflicting definitions across regions).
This enables practical routing: “Tough Calls” (aleatoric) are routed to senior reviewers, while “Policy Gaps” (epistemic) trigger policy updates or model retraining with active learning. In doing so, LPPs do not merely indicate uncertainty; they also shorten the policy-improvement loop and focus scarce human effort, thereby minimizing the cost of the entire moderation system.

We evaluate the LPPs across multiple families of LLMs, off-the-shelf (Gemini, GPT) and open-source (Llama, Qwen), on two moderation datasets, featuring multimodal and multilingual content across multiple risk categories: hate, violence, and more. Across settings, LPPs consistently improve the accuracy and cost frontiers relative to each existing uncertainty estimator. We release reproducibility code alongside the paper to enable practitioners to build upon this framework and facilitate future research on LLM Performance Prediction.\footnote{\url{https://github.com/ZEFR-INC/lpp-research}}

\myparagraph{Contributions} 
(1) We propose a novel framework for estimating LLM uncertainty, learning a meta-model based on LLM Performance Predictors (LPPs) derived from the LLM's output; (2) we demonstrate the merits of our method for selective escalation in a real-world hybrid human-AI workflow for content moderation compared to existing uncertainty estimators; and (3) we introduce novel moderation-oriented uncertainty attribution indicators and show their contribution to explainability, offering insights into when and why LLMs fail, thereby improving learning efficiency in human-AI workflows.

\section{Related Work}
Content moderation refers to the process of detecting user (e.g.,~\cite{gorwa+al:20, kumar+al:24, kolla+al:24}) and AI (e.g.,~\cite{zellers+al:19, lloyd+al:23, fisher+al:24}) generated content that violates laws or policies (cf.~\cite{gorwa+al:20, grimmelmann:15}). The rapid growth of user-generated content, coupled with the vast and evolving range of moderation topics across tasks and use cases, renders manual review impractical, particularly in production environments. This situation drives the increased adoption of unified LLM-based systems for scalable moderation~\cite{kumar+al:24, kolla+al:24, huang:25, levi+al:25}. Recent surveys highlight that uncertainty quantification is a central challenge for deploying such systems reliably at scale~\cite{shorinwa2025surveyuncertaintyquantificationlarge}, and new methods emphasize efficient estimation techniques suitable for production use~\cite{efficient-and-effective}.

The use of LLMs in production moderation systems critically depends on trustworthy confidence estimates, as well-calibrated confidence scores enable safe automation (e.g., selective deferral to human reviewers~\cite{jung+al:24, mozannar2021consistentestimatorslearningdefer}). There is a large body of work on uncertainty quantification in LLMs~\cite{tian+al:23, kadavath+al:22, lin+al:22, kuhn+:23, xiong+al:23, jiang+al:21, manakul+al:23, zhang+al:22, shorinwa2025surveyuncertaintyquantificationlarge,barshalom2025tokenprobabilitieslearnablefast, watson2023explaining, kendall+al:17, farr+al:25, kapoor2024large, cohen2024i, efficient-and-effective, liu+al:24}. Prior work has examined verbalized confidence~\cite{tian+al:23, kadavath+al:22, lin+al:22, xiong+al:23}, but empirical studies consistently find such methods poorly calibrated and highly sensitive to prompting and model type~\cite{kuhn+:23, xiong+al:23}. In addition, multi-sample methods are computationally expensive at production scale. These findings demonstrate that a single, universal uncertainty metric is unreliable and motivate a more robust, multifaceted approach for model failure prediction. To address this, alternative strategies have been explored, including logit-based calibration~\cite{jiang+al:21}, reflection or self-consistency signals~\cite{manakul+al:23}, and reinforcement prompt learning~\cite{zhang+al:22}. Some of them, particularly logit-based information, are used in this work. 

In this work, we propose a novel framework that leverages a diverse set of predictors to improve both calibration and cost-aware decision-making in moderation. Our feature set fuses logit-based features shown to be effective for textual calibration~\cite{jiang+al:21} with additional predictors—including multimodal and uncertainty-attribution signals—that extend applicability beyond text-only settings. Our framework is evaluated across diverse closed- and open-source LLMs on two complementary moderation datasets, including a multimodal benchmark~\cite{levi+al:25}, and employs a learned classifier to produce calibrated confidence estimates.

Another line of work closely related to ours is query performance predictors (QPP) for ad hoc document retrieval~\cite{carmel+Elad:10}. These methods were devised to predict the effectiveness of search results in the absence of relevance judgments. It is common to divide them into two groups: (i) pre-retrieval predictors, which operate prior to retrieval time and often utilize corpus statistics~\cite{hauff+al:08}, and (ii) post-retrieval predictors, which leverage information from the retrieved list~\cite{shtok+al:16,shtok+al:12,tao+al:14,zhou+croft:07}. 
Recently, and with the advance of LLMs, a new category has emerged: post-generation predictors, which utilize information extracted after LLM generation (e.g., next-token distribution~\cite{huly+al:25}). Pre-retrieval, post-retrieval, and post-generation predictors have also been used to estimate the effectiveness of RAG-based LLM systems~\cite{huly+al:25}. Motivated by this line of work, we incorporate both post-retrieval and post-generation predictors into our feature set, extending them with moderation-oriented attribution features for cost-aware escalation.

\section{Method}
\label{sec:methods}
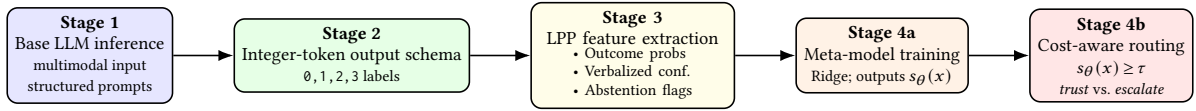
\begin{figure*}[t]
\centering
\begin{tikzpicture}[
  font=\footnotesize,
  node distance=10mm and 8mm, 
  box/.style={draw, rounded corners, semithick, align=center, fill=#1!10},
  arr/.style={-Latex, semithick}
]

\node[box=blue] (s1) {%
  \textbf{Stage 1}\\
  Base LLM inference\\
  \scriptsize multimodal input\\
  \scriptsize structured prompts
};

\node[box=green, right=of s1] (s2) {%
  \textbf{Stage 2}\\
  Integer-token output schema\\
  \scriptsize \texttt{0,1,2,3} labels
};

\node[box=yellow, right=of s2, text width=0.14\textwidth] (s3) {%
  \centering\textbf{Stage 3}\\
  \centering LPP feature extraction\\[-0.25ex]
  \raggedright\scriptsize
  \begin{tabular}{@{}l@{}}
    \textbullet\ Outcome probs\\
    \textbullet\ Verbalized conf.\\
    \textbullet\ Abstention flags
  \end{tabular}
};

\node[box=orange, right=of s3] (s4a) {%
  \textbf{Stage 4a}\\
  Meta-model training\\
  \scriptsize Ridge; outputs $s_\theta(x)$
};

\node[box=red, right=of s4a] (s4b) {%
  \textbf{Stage 4b}\\
  Cost-aware routing\\
  \scriptsize $s_\theta(x)\!\ge\!\tau$\\
  \scriptsize \emph{trust} vs. \emph{escalate}
};

\draw[arr] (s1) -- (s2);
\draw[arr] (s2) -- (s3);
\draw[arr] (s3) -- (s4a);
\draw[arr] (s4a) -- (s4b);

\end{tikzpicture}

\caption{A four-stage framework:
\textbf{(1)} Base LLM Inference;
\textbf{(2)} Integer-Token Output Schema;
\textbf{(3)} LPP Feature Extraction;
\textbf{(4a)} Meta-Model Training to estimate $s_\theta(x)$;
\textbf{(4b)} Cost-Aware Routing to \emph{trust} or \emph{escalate}.}
\label{fig:pipeline}
\end{figure*}

Our method addresses the core challenge of uncertainty-aware human-AI collaboration through a principled four-stage framework. Given the increasingly critical role of LLMs in high-stakes content moderation, we propose a supervised approach to uncertainty quantification that learns when to trust model outputs versus when to escalate for human review. This design directly addresses our research questions: \textbf{RQ1} (can we accurately predict LLM errors?) and \textbf{RQ2} (can selective escalation based on these predictions reduce the overall cost in human-AI moderation workflows?).
As illustrated in Figure~\ref{fig:pipeline}, our framework proceeds in four stages:
(1) \firstmention{Base LLM Inference}: multimodal content is processed by a base LLM using structured prompts that enforce deterministic, token-aligned outputs; (2) \firstmention{Integer-Token Output Schema}: to ensure consistent probability extraction across diverse base LLMs, we designed prompts that constrain outputs to single integer tokens: \texttt{0}~=~``no'', \texttt{1}~=~``yes'', \texttt{2}~=~``inconclusive\_evidence'' (aleatoric uncertainty), \texttt{and 3}~=~``inconclusive\_definition'' (epistemic uncertainty), which are validated against a fixed schema; invalid (malformed) outputs are retried deterministically until a valid integer is obtained ($\leq 3$ retries). These tokens are the basis for uncertainty feature computations~\cite{kapoor2024large,farr-etal-2025-red,gupta2024language,orgad2025llmsknowshowintrinsic}.\footnote{Prior work has also proposed an alternative strategy of adding a dedicated uncertainty token (e.g., ``[IDK]''), enabling the model to explicitly allocate probability mass to ``I don’t know'' rather than forcing a distribution over fixed labels~\cite{cohen2024i}.} Reasoning tokens are handled separately: they remain free-form and are used only for deriving complementary reasoning-based features (e.g., perplexity, per-token entropy), not for outcome probability computation~\cite{guerreiro-etal-2023-looking,efficient-and-effective}; (3) \firstmention{LPP Feature Extraction}: a comprehensive set of LLM Performance Predictors (LPPs) is computed from model outputs, leveraging {\em gray-box} access (token-level log-probabilities, entropy, and reasoning-path statistics) together with {\em black-box} compatible features (Verbalized Confidence and Uncertainty Attribution Indicators; see Section~\ref{subsec:lpps}); (4a) \firstmention{Meta-Model Training}: a Ridge Regression classifier is trained on LPPs to predict LLM correctness, providing calibrated uncertainty estimates~\cite{arthur+al:00}; and (4b) \firstmention{Cost-Aware Routing}: a threshold-based policy decides whether to trust the LLM prediction or escalate to human review, optimizing operational costs.
This architecture is inspired by Query Performance Prediction (QPP) in Information Retrieval~\cite{carmel+Elad:10}, adapted to the agent-based setting where a meta-classifier acts as a gating agent, coordinating between an LLM agent and human reviewers.
\subsection{LLM Performance Predictors (LPPs)}
\label{subsec:lpps}
Our LLM Performance Predictors (LPPs) are a comprehensive feature set primarily extracted via \textbf{gray-box access}—requiring token-level log-probabilities and structured outputs. This feature set also incorporates black-box compatible features (Verbalized Confidence and Uncertainty Attribution Indicators) derived solely from the structured text output. This strategy ensures broad applicability: all evaluated base LLMs provide the necessary gray-box and black-box signals. The combination of gray-box and black-box features strikes a practical balance between information richness and scalability, avoiding reliance on internal activations (white-box) while providing richer signals than output-only methods. Recent analysis has shown that probabilistic confidence (derived from token probabilities) and verbalized confidence capture complementary aspects of model uncertainty, with the former being more accurate but requiring threshold calibration, while the latter provides reasonable signal without additional setup~\cite{ni+al:25}.
LPPs span four families, summarized in Table~\ref{tab:lpps}, and the full mathematical specification of all LPPs appears in Table~\ref{app:full_lpps} (Appendix~\ref{app:prompts}). 
\myparagraph{Post-Hoc Classification Uncertainty}
Features derived from the probability distribution over outcome tokens, computed over the top-$k$ most probable tokens ($k=5$), including Top-5 Entropy $H(\tilde{p})$, Normalized Top-5 Entropy, Effective Choices ($2^{H(\tilde{p})}$), Max Softmax Probability, and Top-2
Probability Margin~\cite{kapoor2024large,kendall+al:17,farr+al:25,watson2023explaining,shorinwa2025surveyuncertaintyquantificationlarge, correa+al:25}.
\myparagraph{Internal Generation Uncertainty}
We implemented Chain-of-Thought (CoT) prompting solely to extract reasoning-sequence features (e.g., Perplexity, Sequence Negative Log-Likelihood, Mean Token Entropy, Token Entropy Quantiles, and Token Probability Quantiles)~\cite{guerreiro-etal-2023-looking,efficient-and-effective,correa+al:25}. 
In our content-moderation setting, CoT inflated confidence and harmed calibration; therefore, these features are documented for completeness but excluded from reported results.
\myparagraph{Verbalized Confidence}
Self-reported confidence extracted from structured outputs: a scalar reported confidence $\hat{c} \in [0,100]$ (normalized to $[0,1]$) and coarse confidence bands (Very Low, Low, Medium, High, Very High) encoded as one-hot features~\cite{kadavath+al:22,tian+al:23,yang2024verbalizedconfidencescoresllms,shorinwa2025surveyuncertaintyquantificationlarge,ni+al:25,xiong+al:23}.
\myparagraph{Uncertainty Attribution Indicators}
Binary indicators when the LLM emits \textsf{inconclusive\_evidence} (aleatoric uncertainty due to missing context, ambiguous frames, etc.) or \textsf{inconclusive\_definition} (epistemic uncertainty due to policy gaps or edge cases). These novel moderation-oriented features enable interpretable routing~\cite{kendall+al:17}: evidence deficient cases escalate to senior reviewers, while policy ambiguous cases trigger guideline updates or active learning cycles.
\begin{table}[h]
\centering
\caption{\textbf{Overview of LPP Feature Families.} Each category captures a distinct aspect of model uncertainty and is accessible via gray-box or black-box access. The full mathematical specification of all LPPs appears in Table~\ref{app:full_lpps} (Appendix~\ref{app:prompts}).}
\small
\setlength{\tabcolsep}{3.5pt}
\renewcommand{\arraystretch}{1.05}
\begin{tabularx}{\columnwidth}{@{}%
  >{\raggedright\arraybackslash}p{0.24\columnwidth}%
  >{\raggedright\arraybackslash}X%
  >{\raggedright\arraybackslash}p{0.3\columnwidth}%
@{}}
\toprule
\textbf{LPP Category} & \textbf{Signal Source \& Intuition} & \textbf{Example Features} \\
\midrule
\textbf{Post-Hoc Classification Uncertainty} (Gray-Box) &
Confidence at the final decision boundary, derived from token log-probabilities over discrete classification outcomes. &
Max Softmax Probability (MSP), Top-5 Entropy (Entropy), Top-2 Probability Margin (Top-2 Margin), Confidence Score \\
\midrule
\textbf{Internal Generation Uncertainty} (Gray-Box; excluded from reported results) &
Signals from intermediate Chain-of-Thought (CoT) reasoning tokens. &
Perplexity (Natural Base), Sequence Negative Log-Likelihood, Mean Token Entropy, Token Entropy Quantiles \\
\midrule
\textbf{Verbalized Confidence} (Black-Box) &
Explicit, self-reported confidence stated by the LLM in structured natural language outputs. &
Reported Confidence (Scalar), Confidence Bands (One-Hot; VL/L/M/H/VH) \\
\midrule
\textbf{Uncertainty Attribution Indicators} (Black-Box) &
Uncertainty attribution indicators signals distinguishing aleatoric uncertainty (evidence deficits) from epistemic uncertainty (policy gaps). Enables interpretable routing. &
Inconclusive Flag (Binary), Evidence-Deficit Indicator, Policy-Gap Indicator \\
\bottomrule
\end{tabularx}
\label{tab:lpps}
\end{table}
\vspace{-11pt}

\subsection{Prompting}
\label{subsec:prompting}
We designed four prompt templates: text-only and multimodal, each with two variants, a direct-answer version (no explicit reasoning steps) and a Chain-of-Thought (CoT) version. All prompts required a structured JSON output with (i) a single integer label (\texttt{0--3}), (ii) an optional reasoning field (populated only in the CoT variant), and (iii) any self-reported confidence. The integer-coded schema aligns labels with token-level log-probabilities. For completeness, we implemented both variants; unless otherwise stated, all results reported in Section~\ref{sec:results} use the direct-answer variant.
\subsection{Cost-Aware Escalation Policy}
\label{subsec:policy}
The meta-model output is a score $s_\theta(x) \in [0,1]$ representing the probability that the base LLM is correct. To operationalize this into a binary trust-or-escalate decision, we employ a \textbf{cost-aware threshold policy}. A prediction is trusted if $s_\theta(x) \geq \tau$; otherwise, it is escalated to human review~\cite{elkan+al:01}.
Let $c_{\text{mis}}$ denote the cost of an undetected misclassification (allowing an LLM error into production) and $c_{\text{rev}}$ the cost of human review. The expected cost, relative to a baseline of always trusting the LLM, is:
\[
\mathcal{C}(\tau) = c_{\text{mis}} \cdot \text{FP} + (c_{\text{rev}} - c_{\text{mis}}) \cdot \text{TN} + c_{\text{rev}} \cdot \text{FN},
\]
where FP, TN, FN are counts of selector decisions (trust/escalate) against LLM correctness (correct/incorrect). Here:
\begin{itemize}
    \item \textbf{TP (True Positives)}: Trusting a correct LLM prediction $\Rightarrow$ no costs.
    \item \textbf{FP (False Positives)}: Trusting an incorrect LLM prediction $\Rightarrow$ misclassification cost (error is missed).
    \item \textbf{TN (True Negatives)}: Escalating an incorrect LLM prediction $\Rightarrow$ review cost offset by avoided misclassification cost (error is caught).
    \item \textbf{FN (False Negatives)}: Escalating a correct LLM prediction $\Rightarrow$ unnecessary review cost.
\end{itemize}
The cost of human reviewers was calculated by multiplying the total review time by their hourly rate (in \$). The cost of misclassification was estimated based on projected business losses, specifically the reduction in customer lifetime value caused by churn due to undetected errors. In our experiments, we set the cost ratio such that $c_{\text{rev}} / c_{\text{mis}} \approx 0.64$, which serves as a reasonable baseline and does not alter relative rankings or cost–accuracy trends under variation.
The operating threshold $\tau^{\star}$ is optimized on a held-out validation set (20\% of the training data) by sweeping $\tau \in [0.35, 0.70]$, a practical range within which the optimum consistently lies near the midpoint (0.5). The value minimizing $\mathcal{C}(\tau)$ is then fixed and applied to the held-out test set for evaluation.
\subsection{Meta-Model Training}
\label{subsec:training}
Our meta-model is a supervised binary classifier trained to predict whether the base LLM is correct ($z=1$) or incorrect ($z=0$) given the extracted LPPs. Based on extensive preliminary experiments, we selected \textbf{Ridge Regression}, whose output scores were converted into probabilities using either sigmoid or isotonic calibration via scikit-learn's \texttt{CalibratedClassifierCV}. This approach was chosen for the following reasons:
\myparagraph{Calibration and Robustness} Ridge Regression's $L_2$ regularization provides crucial protection against multicollinearity, expected within the LPP feature set where signals like entropy and max probability are inherently correlated. This regularization prevents overfitting and improves out-of-sample calibration, critical for reliable uncertainty quantification in production systems.
\myparagraph{Interpretability} Ridge coefficients can be inspected to understand which LPPs are most predictive of errors, supporting explainability requirements in high-stakes moderation workflows.
\myparagraph{Gray-Box Applicability} Ridge Regression does not require model internals (hidden states, gradients), ensuring compatibility with both off-the-shelf APIs and open-source LLMs.
\myparagraph{Handling Class Imbalance}
In the Multimodal Moderation Dataset (1,500 samples), base LLM predictions were correct in roughly 80--90\% of cases (10--20\% errors), resulting in a strong class imbalance. To address this, we (1) downsampled the majority class. We applied a hybrid undersampling strategy combining Tomek Links~\cite{wonjae+al:22} for boundary cleaning with random undersampling of the remaining majority samples. This reduces redundancy while maintaining representative class ratios, and ensures that rare abstention cases ($<5\%$ of data) are always retained. We also tested oversampling methods such as SMOTE~\cite{chawla+al:02}, which yielded comparable performance; for simplicity and reproducibility, we report results using the Tomek+random undersampling approach. We also tuned Ridge’s \texttt{class\_weight} via grid search, testing ratios informed by the cost structure: $w_0 / w_1 \approx c_{\text{rev}} / c_{\text{mis}} \approx 0.64$.
After downsampling and stratified train–test splitting, our typical training set contains $\sim$800 samples and the test set $\sim$300 samples for the Multimodal Moderation Dataset, and $\sim$3{,}500 training and $\sim$900 test samples for the OpenAI Moderation Dataset, with both classes well represented. To ensure fair comparison across LLMs under class imbalance, we fixed the number of negative (error) cases in the test sets: 45 for the Multimodal Moderation Dataset and 150 for the OpenAI Moderation Dataset. This controlled evaluation design mitigates variability due to scarcity of negative samples while preserving comparability across models.
\myparagraph{Cross-Validation and Hyperparameter Search}
We employ stratified $3$-fold cross-validation and a comprehensive grid search on the training portion of the data (after holding out a validation split for threshold selection) over:
\begin{itemize}
    \item \texttt{alpha} $\in \{0.1, 1.0, 10.0, 100.0\}$ (regularization strength)
    \item \texttt{solver} $\in \{\text{auto}, \text{lsqr}\}$
    \item \texttt{tol} $\in \{10^{-6}, 10^{-5}, 10^{-4}, 10^{-3}\}$ (convergence tolerance)
    \item \texttt{max\_iter} $\in \{1000, 2000, 3000\}$
    \item \texttt{class\_weight}: 7 configurations including cost-informed ratios and ``balanced''
    \item Calibration \texttt{method} $\in \{\text{sigmoid, isotonic}\}$
\end{itemize}
Models are selected based on F1-score on the minority class (errors), as this metric balances precision and recall for the operationally critical decision of identifying when the LLM is wrong.\\
Formally, each sample is labeled with a binary correctness indicator $z_i \in \{0,1\}$, equal to $1$ if the LLM prediction matches the ground truth and $0$ otherwise.  
The ridge regression objective is
\[
\min_{w,b} \; \frac{1}{N}\sum_{i=1}^N \big(z_i - (w^\top f(x_i) + b)\big)^2 + \lambda \|w\|_2^2,
\]
where $f(x_i) \in \mathbb{R}^d$ is the feature vector, $w \in \mathbb{R}^d$ and $b \in \mathbb{R}$ are model parameters, and $\lambda \ge 0$ controls $L_2$ regularization. This yields a calibrated linear predictor of correctness probability, with isotonic or Platt scaling to map scores into $[0,1]$.
Inference settings (including deterministic decoding) are described in Section~\ref{subsec:experimental_setup} and Appendix~\ref{app:prompts}.
\section{Experiments}
\label{sec:experiments}
We evaluate the proposed LLM Performance Predictor (LPP) framework in the context of cost-aware selective escalation for human–AI content moderation. From a multi-agent systems perspective, our experiments study the coordination dynamics between autonomous LLM agents and human reviewers, where the LPP meta-model serves as a decision-making intermediary that allocates tasks under cost constraints.
\subsection{Experimental Setup}
\label{subsec:experimental_setup}
We evaluate several LLMs: proprietary (gpt-4o-mini, gpt-4o, gpt-4.1-mini, gemini-2.5-flash-lite, gemini-2.0-flash-lite, gemini-2.0-flash-001) and open-source (QWEN3-14B, QWEN3-32B, LLAMA32-11B).
\footnote{We considered including safety-specialized models such as Llama Guard~\cite{Inan2023LlamaGL}, but defer discussion of these to Section~\ref{subsec:limitations}.}
Following Section~\ref{sec:methods}, we instantiate LPPs using ridge regression trained on heterogeneous feature families derived from intrinsic model signals (e.g., token-level probabilities, self-consistency markers) and post-hoc signals (e.g., calibration-based scores). A summary of feature groups is provided in Table~\ref{tab:lpps}. Hyperparameters for baselines and the meta-model are tuned via nested cross-validation, with stratified sampling. To address class imbalance, we apply down-sampling as described in Sec.~\ref{subsec:training}, and evaluate final performance on a held-out test set using the optimized threshold.
The meta-model is trained with Ridge Regression, chosen for its robustness in high-dimensional feature spaces.

\myparagraph{Additional Experimental Details}
We use deterministic decoding parameters (\texttt{temperature}=0, \texttt{top-p}=1) with a fixed random seed (\texttt{random\_state}=42) for data splitting and cross-validation reproducibility. For each generated token, we request log-probabilities of the \textit{top-20} candidate tokens. From these we compute two feature families:
\begin{enumerate}
    \item \textbf{Unfiltered top-5 features}: standard uncertainty metrics computed over the five most probable tokens.
    \item \textbf{Filtered features}: log-probabilities restricted to the four valid class labels (\texttt{0-3}).
\end{enumerate}
This dual representation captures both the model’s general uncertainty landscape and its calibrated confidence over the task’s decision space. CoT perplexity is computed with natural logarithms. Cross-validation uses \texttt{StratifiedKFold} with $K=3$ splits.

\subsection{Datasets}
We use two complementary datasets:  
(i) the \textbf{OpenAI Moderation dataset}~\cite{markov2023holisticapproachundesiredcontent}, which contains 1,680 English text samples annotated across multiple moderation categories (hate, self-harm, sexual, and violence, with fine-grained subcategories such as hate/threatening, sexual/minors, and violence/graphic). Since the dataset is multi-label, we evaluate our model at the level of \emph{text–label pairs}: each annotation is treated as a separate prediction instance. Thus, while the corpus comprises 1,680 source texts, the effective number of evaluation instances is larger, reflecting the total set of text–label assignments. This framing allows a more precise assessment of category-specific moderation performance.  
(ii) a \textbf{Multimodal Moderation Dataset}~\cite{levi+al:25} of $1{,}500$ short-form videos spanning three categories: Death, Injury and Military Conflict; Drugs, Alcohol and Tobacco; and Kids content, across 12 languages and four modalities (text, thumbnail, transcript, video/frame). Together, these datasets enable evaluation of both text-only and multimodal moderation scenarios.  
\begin{table}[t]
\centering
\caption{\textbf{Summary of datasets.}}
\small
\setlength{\tabcolsep}{3.5pt}
\renewcommand{\arraystretch}{1.05}
\begin{tabularx}{\columnwidth}{@{}%
  >{\raggedright\arraybackslash}p{0.22\columnwidth}%
  >{\raggedright\arraybackslash}p{0.12\columnwidth}%
  >{\raggedright\arraybackslash}p{0.22\columnwidth}%
  >{\raggedright\arraybackslash}X%
@{}}
\toprule
\textbf{Dataset} & \textbf{Size} & \textbf{Languages / Modalities} & \textbf{Categories} \\
\midrule
OpenAI Moderation~\cite{markov2023holisticapproachundesiredcontent} &
1{,}680 texts &
English / Text &
Hate, Self-harm, Sexual, Violence (with subcategories) \\
\midrule
Multimodal Moderation~\cite{levi+al:25} &
1{,}500 videos &
12 languages / Text, Thumbnail, Transcript, Video/Frames &
DIMC (Death/Injury/Military Conflict), DAT (Drugs/Alcohol/Tobacco), Kids \\
\bottomrule
\end{tabularx}
\label{tab:datasets}
\end{table}
\subsection{Baselines}
We compare our LPP-based meta-model against widely used uncertainty heuristics:
\begin{itemize}
    \item \textbf{MSP}~\cite{hendrycks2017a,geifman2017selectiveclassificationdeepneural}.
    \item \textbf{Top-2 Margin}~\cite{gupta2024language}.
    \item \textbf{Entropy}~\cite{kendall+al:17,barshalom2025tokenprobabilitieslearnablefast}.
\end{itemize}
These represent the strongest post-hoc uncertainty signals. We also include a cost-insensitive \emph{always-trust} baseline.
\subsection{Evaluation Metrics}
To address \textbf{RQ1}, we evaluate error prediction using:
\begin{itemize}
    \item \textbf{AUC-ROC}: area under the receiver operating characteristic curve, measuring ranking quality of correct vs.\ incorrect predictions across thresholds.
    \item \textbf{F1-Score}: harmonic mean of precision and recall, balancing false positives and false negatives.
    \item \textbf{Macro-F1}: F1 macro-averaged over correctness labels (correct vs.\ incorrect), giving equal weight to error and non-error cases.
\end{itemize}
To address \textbf{RQ2}, we evaluate:
\begin{itemize}
    \item \textbf{Expected Cost $\mathcal{C}(\tau)$}, as defined in Sec.~\ref{subsec:policy}
    \item \textbf{Escalation Ratio}, defined as the fraction of items routed to human review:
    \[
    \text{Escalation Ratio} = \frac{TN + FN}{TP + FP + TN + FN}.
    \]
    Following best practice in selective prediction~\cite{geifman2017selectiveclassificationdeepneural,hendrycks2017a}, we report both the \emph{percentage} of items escalated and the \emph{absolute number of escalations}.
\end{itemize}
\section{Results}
\label{sec:results}
Our meta-model outperforms standard uncertainty estimators in predictive accuracy and cost-aware decision-making across datasets and LLM families. Although we implemented CoT variants, all reported results use direct-answer prompting, as CoT increased confidence without improving calibration.
\subsection{Main Performance Benchmarks: Predictive Accuracy (RQ1)}
\label{subsec:predictive_performance}
Tables~\ref{tab:main_results_public_no_cot} and~\ref{tab:main_results_holdout_no_cot} report F1, AUC-ROC, and Macro-F1 for error prediction across nine LLMs overall, evaluated on two datasets: OpenAI Moderation Dataset~\cite{markov2023holisticapproachundesiredcontent} and a Multimodal Moderation Dataset~\cite{levi+al:25}. The meta-model consistently exceeds MSP, Top-2 Margin, and Entropy baselines.

\begin{table*}[t]
\centering
\caption{Baseline vs.\ meta-model predictive performance on the OpenAI Moderation Dataset. Metrics are reported as percentages. Bold values indicate the best result per metric for each LLM (row-wise).}
\label{tab:main_results_public_no_cot}
\resizebox{\linewidth}{!}{
\begin{tabular}{lcccccccccccc}
\toprule
 & \multicolumn{3}{c}{MSP} & \multicolumn{3}{c}{Top-2 Margin} & \multicolumn{3}{c}{Entropy} & \multicolumn{3}{c}{Meta-Model (Ours)} \\
\cmidrule(lr){2-4} \cmidrule(lr){5-7} \cmidrule(lr){8-10} \cmidrule(lr){11-13}
LLM & F1 & AUC-ROC & Macro-F1 & F1 & AUC-ROC & Macro-F1 & F1 & AUC-ROC & Macro-F1 & F1 & AUC-ROC & Macro-F1 \\
\midrule
gpt-4o-mini         & 81.27\% & 83.55\% & 64.96\% & 81.27\% & 83.55\% & 64.96\% & 80.55\% & 83.43\% & 64.20\% & \textbf{94.14\%} & \textbf{93.46\%} & \textbf{82.31\%} \\
gpt-4o              & 84.35\% & 90.34\% & 71.14\% & 84.41\% & 90.33\% & 70.96\% & 84.35\% & \textbf{90.48\%} & 71.14\% & \textbf{91.35\%} & 90.64\% & \textbf{76.81\%} \\
gpt-4.1-mini        & 88.79\% & 91.10\% & 75.28\% & 88.21\% & 91.04\% & 74.56\% & 88.71\% & 91.07\% & 75.16\% & \textbf{91.93\%} & \textbf{91.73\%} & \textbf{79.94\%} \\
gemini-2.5-flash-lite & 81.04\% & 81.69\% & 65.67\% & 82.70\% & 81.41\% & 66.51\% & 80.95\% & 82.22\% & 65.58\% & \textbf{89.11\%} & \textbf{87.58\%} & \textbf{74.00\%} \\
gemini-2.0-flash-lite & 88.73\% & 85.57\% & 74.52\% & 88.48\% & 85.66\% & 74.26\% & 88.73\% & 85.52\% & 74.52\% & \textbf{93.55\%} & \textbf{89.76\%} & \textbf{81.39\%} \\
gemini-2.0-flash-001  & 85.29\% & 82.76\% & 70.39\% & 85.29\% & 82.76\% & 70.39\% & 85.29\% & 82.76\% & 70.39\% & \textbf{89.38\%} & \textbf{86.34\%} & \textbf{75.36\%} \\
QWEN3-14B  & 90.93\% & 89.94\% & 64.25\% & 90.88\% & 89.90\% & 63.54\% & \textbf{91.07}\% & 89.99\% & 66.52\% & 90.54\% & \textbf{92.07}\% & \textbf{77.20}\% \\

\end{tabular}}
\end{table*}
\begin{table*}[t]
\centering
\caption{Baseline vs.\ meta-model predictive performance on the Multimodal Moderation Dataset. Metrics are reported as percentages. Bold values indicate the best result per metric for each LLM (row-wise).}
\label{tab:main_results_holdout_no_cot}
\resizebox{\linewidth}{!}{
\begin{tabular}{lcccccccccccc}
\toprule
 & \multicolumn{3}{c}{MSP} & \multicolumn{3}{c}{Top-2 Margin} & \multicolumn{3}{c}{Entropy} & \multicolumn{3}{c}{Meta-Model (Ours)} \\
\cmidrule(lr){2-4} \cmidrule(lr){5-7} \cmidrule(lr){8-10} \cmidrule(lr){11-13}
LLM & F1 & AUC-ROC & Macro-F1 & F1 & AUC-ROC & Macro-F1 & F1 & AUC-ROC & Macro-F1 & F1 & AUC-ROC & Macro-F1 \\
\midrule
gpt-4o-mini         & 85.71\% & 83.87\% & 71.20\% & 85.71\% & 83.95\% & 71.20\% & 85.12\% & 83.69\% & 70.47\% & \textbf{87.34\%} & \textbf{88.71\%} & \textbf{74.07\%} \\
gpt-4o              & 88.05\% & 84.97\% & 67.39\% & 87.58\% & 84.72\% & 66.73\% & 88.00\% & \textbf{84.98\%} & 67.85\% & \textbf{91.42\%} & 84.43\% & \textbf{69.80\%} \\
gpt-4.1-mini        & 88.84\% & \textbf{84.93\%} & \textbf{72.39\%} & 88.84\% & 84.82\% & 72.39\% & 88.84\% & 84.93\% & 72.39\% & \textbf{90.98\%} & 84.18\% & 69.02\% \\
gemini-2.5-flash-lite & 84.53\% & 75.18\% & 63.87\% & 85.28\% & \textbf{75.20\%} & \textbf{64.77\%} & 83.19\% & 75.21\% & 62.81\% & \textbf{90.56\%} & 68.26\% & 57.59\% \\
gemini-2.0-flash-lite & 69.52\% & 67.34\% & 52.41\% & 69.85\% & 67.29\% & 52.67\% & 69.52\% & 67.37\% & 52.41\% & \textbf{85.47\%} & \textbf{69.37\%} & \textbf{61.09\%} \\
gemini-2.0-flash-001      & \textbf{91.65\%} & 62.38\% & 45.82\% & 91.30\% & 62.52\% & 52.17\% & 91.65\% & 62.32\% & 45.82\% & 90.71\% & \textbf{66.37\%} & \textbf{56.47\%} \\
QWEN3-14B            & \textbf{91.06\%} & 74.04\% & 60.04\% & 91.06\% & 73.92\% & 60.04\% & 90.57\% & \textbf{74.25\%} & 60.44\% & 86.49\% & 73.77\% & \textbf{65.97\%} \\
QWEN3-32B            & \textbf{87.70\%} & 74.70\% & 55.10\% & 87.70\% & 74.52\% & 55.10\% & 87.25\% & \textbf{74.79\%} & 54.60\% & 86.82\% & 74.71\% & \textbf{57.69\%} \\
LLAMA32-11B          & \textbf{86.32\%} & \textbf{70.79\%} & 56.02\% & 86.32\% & 70.49\% & 56.02\% & 86.02\% & 70.74\% & 59.68\% & 79.51\% & 70.67\% & \textbf{62.52\%} \\
\bottomrule
\end{tabular}}
\end{table*}

\myparagraph{Text-Only (OpenAI Moderation Dataset)}
The meta-model improves ranking and class balance across models. Relative to the strongest baseline in each row, for \textbf{gpt-4o-mini}, F1 increases from 81.27\% to 94.14\%, AUC-ROC from 83.55\% to 93.46\%, and Macro-F1 from 64.96\% to 82.31\%. Similar improvements are observed for \textbf{gemini-2.5-flash-lite}, where F1 increases from 82.70\% to 89.11\%, AUC-ROC from 82.22\% to 87.58\%, and Macro-F1 from 66.51\% to 74.00\%, and for \textbf{gemini-2.0-flash-lite}, where F1 increases from 88.73\% to 93.55\%, AUC-ROC from 85.66\% to 89.76\%, and Macro-F1 from 74.52\% to 81.39\%. For the stronger-performing models \textbf{gpt-4.1-mini} and \textbf{gpt-4o}, F1 increases from 88.79\% to 91.93\% and from 84.41\% to 91.35\%, respectively. Entries for \textbf{LLAMA32-11B} and \textbf{QWEN3-32B} are omitted from the OpenAI Moderation Dataset results. The former could not be reliably evaluated due to challenges in parsing its outputs across the dataset’s taxonomy, while the latter required compute resources beyond the scope of our evaluation. These omissions do not affect the aggregate trends reported.

\myparagraph{Multimodal (Multimodal Moderation Dataset)}
On the harder Multimodal Moderation Dataset, effects are more model-dependent. For \textbf{gpt-4o-mini}, F1 increases from 85.71\% to 87.34\%, AUC-ROC from 83.95\% to 88.71\%, and Macro-F1 from 71.20\% to 74.07\%. For \textbf{gpt-4o}, F1 increases from 88.05\% to 91.42\%, while Macro-F1 increases from 67.85\% to 69.80\%. The largest gains are observed for \textbf{gemini-2.0-flash-lite}, where F1 increases from 69.85\% to 85.47\% and Macro-F1 from 52.67\% to 61.09\%. In contrast, some models exhibit trade-offs: for \textbf{gpt-4.1-mini}, Macro-F1 decreases from 72.39\% to 69.02\%, and for \textbf{gemini-2.5-flash-lite}, Macro-F1 decreases from 64.77\% to 57.59\%. In several cases (e.g., \textbf{QWEN3-14B}, \textbf{LLAMA32-11B}), F1 decreases despite gains in Macro-F1, indicating calibration challenges and majority-class bias under distribution shift.

\subsection{Cost-Aware Escalation: Operational Efficiency (RQ2)}
\label{subsec:cost_analysis}
Table~\ref{tab:cost_merged_side_by_side} reports expected costs and escalation rates under the cost-aware policy described in Section~\ref{subsec:policy}. These results address RQ2 by quantifying the economic value of selective escalation in production human--AI workflows.
\begin{table*}[t]
\centering
\caption{Cost-aware evaluation. Metrics: Always-Trust (A.T.) cost (\$), expected cost (\$), escalations (count), and escalation ratio (\%). Bold values indicate the lowest expected cost among methods for each LLM (row-wise).}
\label{tab:cost_merged_side_by_side}
\begin{tabular}{@{}c @{\ \vrule width 1pt\ } c@{}}  
\begin{subtable}[t]{0.56\linewidth}
\centering
\caption{OpenAI Moderation Dataset}
\label{tab:cost_results_public}
\setlength{\tabcolsep}{2.05pt}
\renewcommand{\arraystretch}{1.05}
\resizebox{\linewidth}{!}{
\begin{tabular}{lccccccccccccc}
\toprule
LLM & A.T. & \multicolumn{3}{c}{MSP} & \multicolumn{3}{c}{Top-2 Margin} & \multicolumn{3}{c}{Entropy} & \multicolumn{3}{c}{Meta-Model (Ours)} \\
 &  & Cost & Esc & Ratio & Cost & Esc & Ratio & Cost & Esc & Ratio & Cost & Esc & Ratio \\
\cmidrule(lr){3-5}\cmidrule(lr){6-8}\cmidrule(lr){9-11}\cmidrule(lr){12-14}
gpt-4o-mini   & 127 & 138   & 339 & 38\% & 132 & 331 & 37\% & 138   & 339 & 38\% & \textbf{38} & 148 & 16\% \\
gpt-4o        & 127 & 74    & 275 & 31\% & 75  & 271 & 30\% & 74    & 275 & 31\% & \textbf{51} & 151 & 17\% \\
gpt-4.1-mini  & 127 & 69    & 258 & 29\% & 73  & 267 & 30\% & 69    & 259 & 29\% & \textbf{45} & 212 & 24\% \\
gemini-2.5-flash-lite & 127 & 128 & 347 & 38\% & 122 & 315 & 35\% & 128 & 348 & 39\% & \textbf{77} & 227 & 25\% \\
gemini-2.0-flash-lite & 127 & 74  & 248 & 28\% & 75  & 253 & 28\% & 74  & 248 & 28\% & \textbf{41} & 162 & 18\% \\
gemini-2.0-flash-001      & 127 & 97  & 297 & 33\% & 97  & 297 & 33\% & 97  & 297 & 33\% & \textbf{69} & 238 & 26\% \\
QWEN3-14B     & 127 &    103   &   73  &   8\%     &   105     &  70   &    8\%    &   98     &  84   &    9\%    &    \textbf{56}   &  205   &   23\%     \\
QWEN3-32B     & NA   & NA       & NA     & NA        & NA        & NA     & NA        & NA        & NA     & NA        & NA       & NA     & NA        \\
LLAMA32-11B   & NA   & NA       & NA     & NA        & NA        & NA     & NA        & NA        & NA     & NA        & NA       & NA     & NA        \\
\bottomrule
\end{tabular}}
\end{subtable}
&
\begin{subtable}[t]{0.44\linewidth}
\centering
\caption{Multimodal Moderation Dataset}
\label{tab:cost_results_holdout}
\setlength{\tabcolsep}{2.05pt}
\renewcommand{\arraystretch}{1.05}
\resizebox{\linewidth}{!}{
\begin{tabular}{ccccccccccccc}
\toprule
A.T. & \multicolumn{3}{c}{MSP} & \multicolumn{3}{c}{Top-2 Margin} & \multicolumn{3}{c}{Entropy} & \multicolumn{3}{c}{Meta-Model (Ours)} \\
 & Cost & Esc & Ratio & Cost & Esc & Ratio & Cost & Esc & Ratio & Cost & Esc & Ratio \\
\cmidrule(lr){2-4}\cmidrule(lr){5-7}\cmidrule(lr){8-10}\cmidrule(lr){11-13}
42 & 39 & 110 & 37\% & 37 & 107 & 36\% & 39 & 110 & 37\% & \textbf{22} & 80 & 27\% \\
42 & 41 & 70  & 23\% & 42 & 73  & 24\% & 40 & 72  & 24\% & \textbf{29} & 38 & 13\% \\
42 & 32 & 85  & 28\% & 32 & 85  & 28\% & 32 & 85  & 28\% & \textbf{30} & 40 & 13\% \\
42 & 53 & 93  & 31\% & 51 & 90  & 30\% & 53 & 92  & 31\% & \textbf{40} & 20 & 7\% \\
\textbf{42} & 80 & 163 & 54\% & 80 & 162 & 54\% & 80 & 163 & 54\% & 45 & 64 & 21\% \\
42 & 35 & 7   & 2\%  & 35 & 7   & 2\%  & 35 & 10  & 3\%  & \textbf{34} & 15 & 5\% \\
42 & \textbf{36} & 17 & 6\% & \textbf{36} & 65 & 22\% & \textbf{36} & 21 & 7\% & \textbf{36} & 65 & 22\% \\
\textbf{42} & 47 & 35 & 12\% & 47 & 46 & 15\% & 49 & 37 & 12\% & 47 & 46 & 15\% \\
42 & 41 & 25 & 8\%  & 39 & 78 & 26\% & \textbf{38} & 33 & 11\% & 39 & 78 & 26\% \\
\bottomrule
\end{tabular}}
\end{subtable}
\\ 
\end{tabular}
\end{table*}
\myparagraph{Significant Cost Reductions on Text-Only Benchmark} On the OpenAI Moderation Dataset, the meta-model achieves substantial cost savings relative to standard baselines for most models. For \textbf{gpt-4o-mini}, expected cost decreases from \$132 (best baseline: Top-2 Margin) to \$38 (71\% reduction), while the number of escalations decreases from 331 to 148. Similarly, for \textbf{gemini-2.5-flash-lite}, expected cost decreases from \$122 to \$77 (36\% reduction) and escalations from 315 to 227. For \textbf{gemini-2.0-flash-lite}, expected cost decreases from \$74 to \$41 (41\% reduction), with escalations decreasing from 248 to 162. These patterns indicate that baseline methods (e.g., MSP, Entropy) tend to over-escalate, triggering unnecessary human reviews that inflate operational costs without proportional gains in error prevention. In contrast, the meta-model’s improved calibration enables more targeted escalation, reducing false alarms while preserving error detection.\footnote{Costs are reported in \$, based on estimated review time and projected business loss per error; exact values are task-dependent and we report relative comparisons.}
\myparagraph{Generalization to Multimodal Workflows} The Multimodal Moderation Dataset exhibits similar trends with notable variation across models. For \textbf{gpt-4o-mini}, expected cost decreases from \$37 (best baseline: Top-2 Margin) to \$22 (41\% reduction), with escalations decreasing from 107 to 80. \textbf{gpt-4o} shows a comparable pattern, with expected cost decreasing from \$42 to \$29 (32\% reduction) and escalations from 73 to 38. Notably, for \textbf{gemini-2.5-flash-lite}, expected cost decreases from \$51 to \$40 (22\% reduction) while the number of escalations decreases from 90 to 20. These results suggest that the meta-model can exploit modality-specific uncertainty signals (e.g., visual ambiguity vs.\ transcript inconsistency) that single-metric baselines fail to capture. However, not all models benefit uniformly: for \textbf{QWEN3-14B} and \textbf{LLAMA32-11B}, expected cost remains comparable to the Top-2 Margin baseline, indicating limited gains from meta-modeling on this dataset.
\subsection{Feature Family Ablations: Decomposing LPP Contributions}
\label{subsec:ablations}
\begin{figure}[t]
    \centering
    \includegraphics[width=0.9\linewidth]{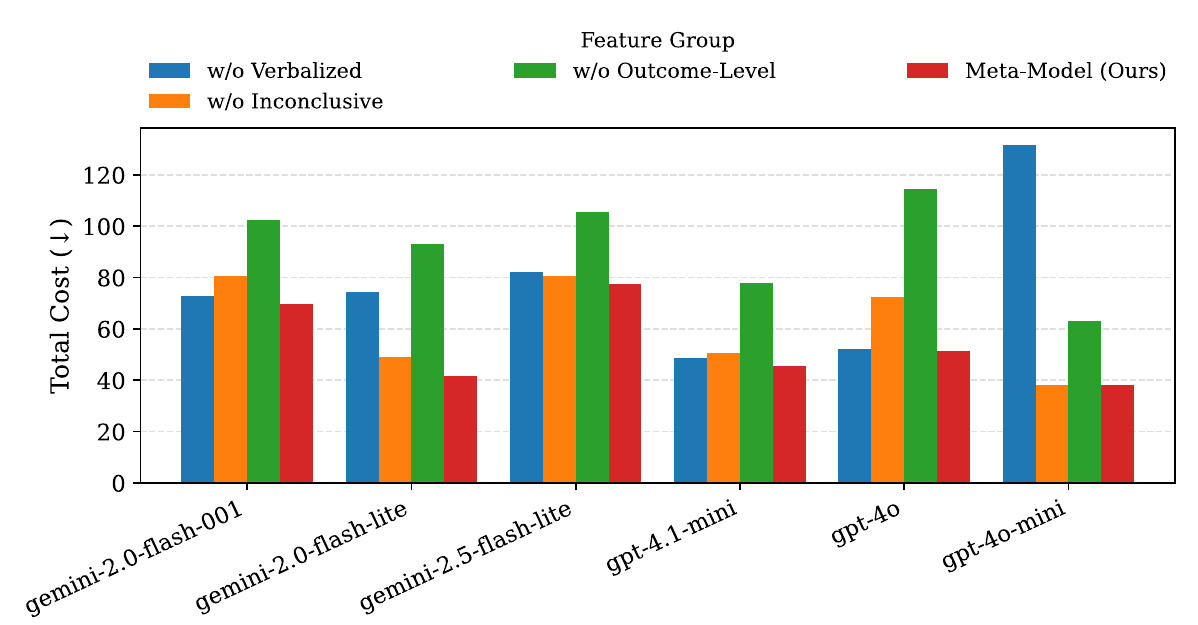}
    \Description{Bar plot of an ablation study on the OpenAI Moderation Dataset with CoT prompting, where each bar shows the total cost after removing one predictor feature family.}
   \caption{Ablation study on the OpenAI Moderation Dataset. Bars show total cost (\$) when removing one feature family, illustrating each predictor group’s contribution.}
    \label{fig:ablation_public_no_cot}
\end{figure}

\begin{figure}[t]
\centering
    \includegraphics[width=0.9\linewidth]{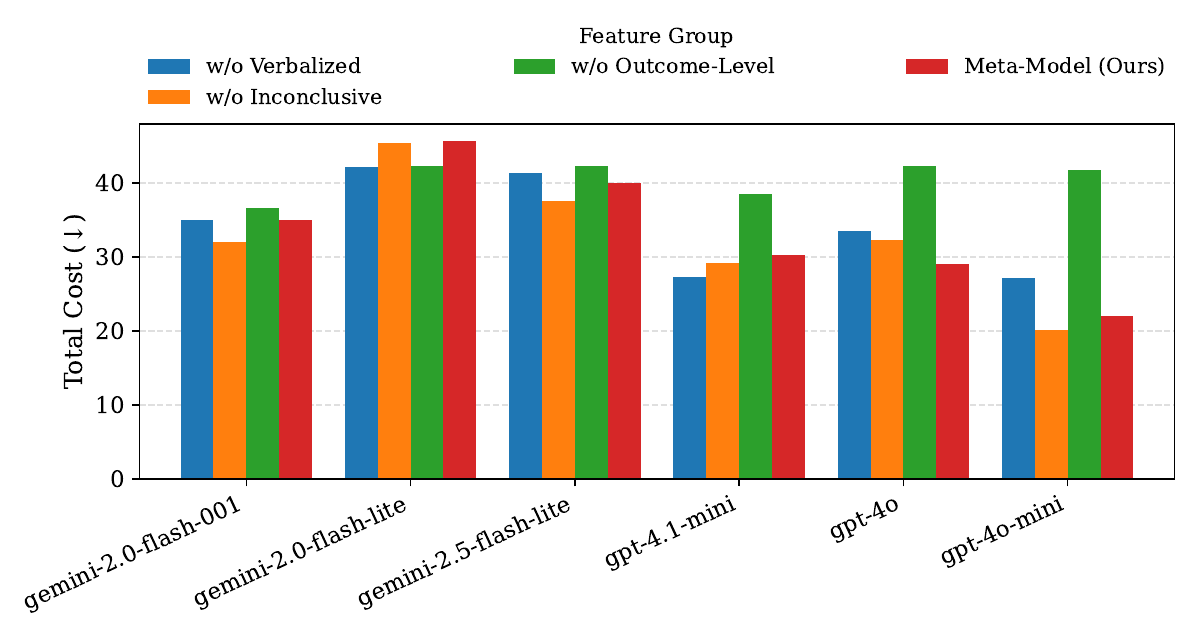}
    \Description{Bar plot of an ablation study on the Multimodal Moderation Dataset without CoT prompting, where each bar shows the total cost after removing one predictor feature family.}
    \caption{Ablation study on the Multimodal Moderation Dataset. Bars show total cost (\$) when removing one feature family, illustrating each predictor group’s contribution.}
    \label{fig:ablation_holdout_no_cot}
    \vspace{-8pt}
\end{figure}

To quantify the individual contributions of each LPP family, we conduct systematic ablation studies where the meta-model is trained with one feature group removed. Figures~\ref{fig:ablation_public_no_cot} and~\ref{fig:ablation_holdout_no_cot} visualize the resulting cost increases relative to the full model across base LLMs.
\myparagraph{Complementary Nature of Feature Families}
Across all settings, removing \emph{any} feature family degrades performance, confirming that the three LPP groups (Post-Hoc Classification Uncertainty, Verbalized Confidence, and Uncertainty Attribution Indicators) capture complementary aspects of model behavior.
In Figures~\ref{fig:ablation_public_no_cot} and~\ref{fig:ablation_holdout_no_cot}, these correspond respectively to \emph{Outcome-Level}, \emph{Verbalized}, and \emph{Inconclusive} feature groups.
On the OpenAI Moderation Dataset (Figure~\ref{fig:ablation_public_no_cot}), removing Post-Hoc Classification Uncertainty features (Entropy, MSP, Top-2 Margin) leads to the largest cost increases for most models, consistent with prior work showing that token-level probabilities are the strongest single uncertainty signal.
Removing verbalized confidence also increases costs by roughly 5--15\%, as indicated by the cost differences in Figures~\ref{fig:ablation_public_no_cot} and~\ref{fig:ablation_holdout_no_cot}, suggesting that self-reported confidence adds complementary signals not fully captured by probabilistic features.

The Uncertainty Attribution Indicators, despite being binary, contribute measurably to cost reduction, with especially visible effects for \textbf{gemini-2.5-flash-lite} and \textbf{gemini-2.0-flash-lite}, where removing them produces clear increases in expected costs. This highlights the value of explicit abstention signals in helping the meta-model avoid false confidence—cases where the base LLM assigns high probability to an incorrect answer due to policy ambiguity or evidence deficits.
\subsection{Cross-Dataset Generalization and Robustness}
\label{subsec:generalization}
A key question for meta-learning is whether patterns transfer across distributions. Our two-dataset evaluation shows \textit{model-dependent} robustness: on the OpenAI Moderation Dataset (text-only), the meta-model consistently outperforms post-hoc baselines, while on the Multimodal Moderation Dataset it improves some models (e.g., gpt-4o-mini, gpt-4o, gemini-2.0-flash-lite) but is matched or outperformed by others (e.g., QWEN, LLAMA), reflecting heterogeneity under domain shift.

Absolute performance naturally degrades for some models when moving from the text-only benchmark to the harder multimodal distribution. For example, for gpt-4.1-mini the Macro-F1 score decreases from \textbf{79.94\%} (text-only) to \textbf{69.02\%}, illustrating the additional complexity introduced by visual ambiguity, multilingual code-switching, and culturally contingent moderation judgments. Despite these shifts, the meta-model retains gains for multiple models on the Multimodal Moderation Dataset, while revealing cases where baseline methods remain competitive, underscoring that generalization depends on both the base LLM family and the evaluation distribution.

These patterns reinforce the \emph{cost-aware} view: even when accuracy gains are mixed under distribution shift, LPP-based escalation yields savings by directing review to the most error-prone cases.
\subsection{Limitations and Open Challenges}
\label{subsec:limitations}
Our study demonstrates substantial progress, but several limitations remain. First, the meta-model requires labeled data samples with ground-truth moderation decisions, which introduces annotation costs. Semi-supervised and active learning strategies could reduce this burden. Second, we focus only on \emph{post-response} LPPs extracted after the LLM generates an answer. Incorporating \emph{pre-response} predictors could enable anticipatory escalation and adaptive coordination.
Third, although we evaluate across a range of LLM families, all are transformer-based autoregressive models. Whether LPPs generalize to other architectures (e.g., retrieval-augmented or neuro-symbolic systems) remains an open question. Fourth, our cost model assumes fixed review and error penalties, whereas in practice costs vary by severity, reviewer expertise, and platform-specific risk tolerance. Extending escalation to a dynamic resource-allocation problem is a promising direction.
Finally, while we extracted Chain-of-Thought (CoT) derived features, we do not advocate CoT prompting in this setting: in our experiments, it inflated confidence and harmed calibration~\cite{tian+al:23,zhang2024studycalibrationincontextlearning}.

More broadly, our experiments focus on content moderation, but the LPP framework is applicable to other high-stakes settings such as fraud detection, compliance, or medical triage, where abstention signals and cost-sensitive escalation are equally critical.
\begin{figure}[t]
    \centering
    \includegraphics[width=0.9\linewidth]{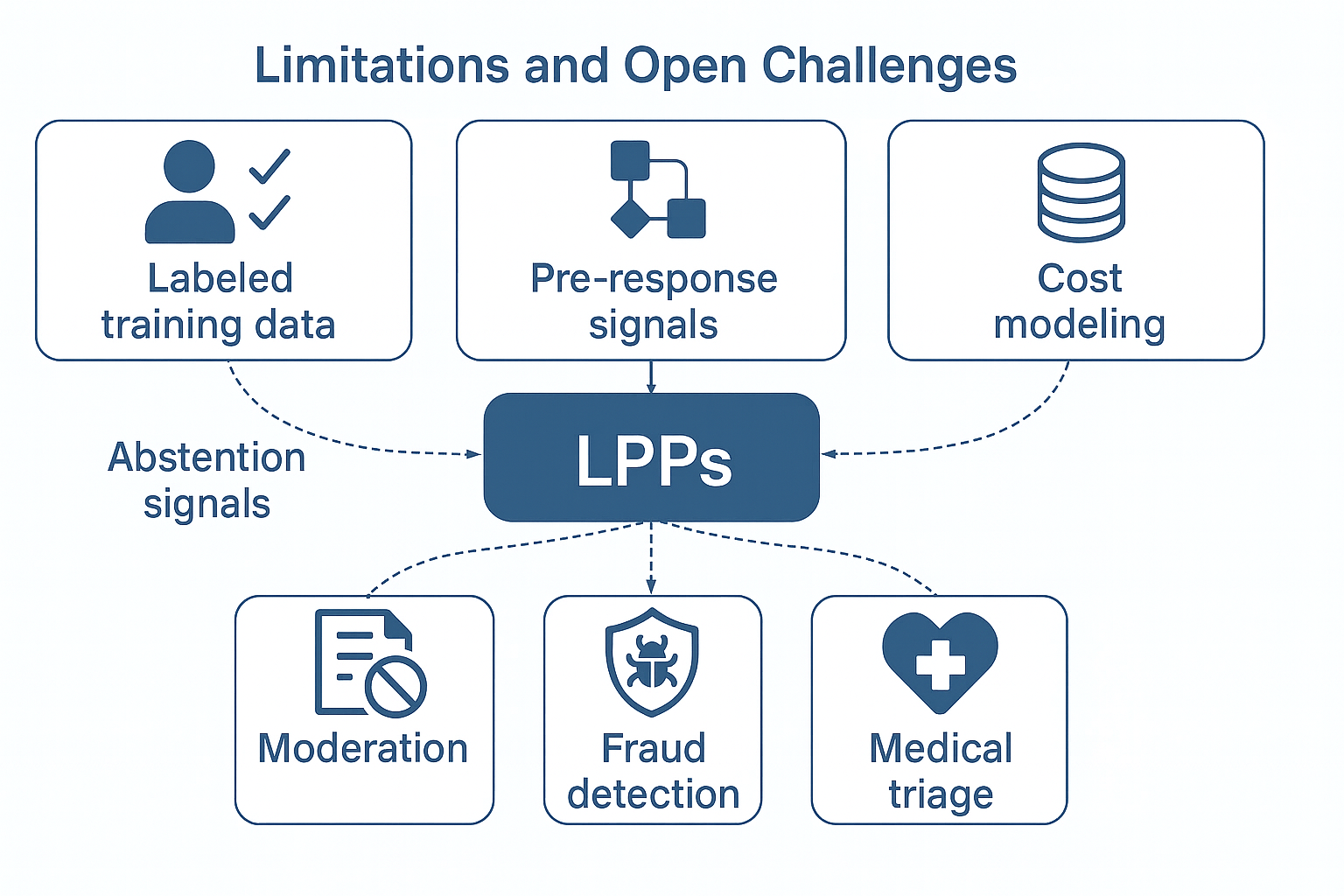}
    \Description{Diagram showing LPPs as the central box connected by arrows to surrounding components. Above, three boxes labeled Labeled training data, Pre-response signals, and Cost modeling feed into LPPs. Below, three boxes labeled Moderation, Fraud detection, and Medical triage receive outputs from LPPs. Dashed arrows indicate abstention signals.}
    \caption{\textbf{LPPs as a cross-domain uncertainty router.} The meta-model routes between automated decisions and human review using uncertainty signals, applicable to moderation, fraud detection, compliance, and medical triage.}
    \label{fig:lpp_router}
    \vspace{-9pt}
\end{figure}
Figure~\ref{fig:lpp_router} illustrates this broader vision: by abstracting uncertainty-aware routing beyond moderation, LPPs can act as a general-purpose coordination mechanism wherever human expertise must be allocated efficiently under uncertainty.
We excluded safety-specialized models like Llama Guard~\cite{Inan2023LlamaGL}, whose taxonomies do not align with our categories (DIMC: Death/Injury/Military Conflict; DAT: Drugs/Alcohol/Tobacco; Kids). Such models may offer stronger abstention signals or calibration, and hybrid setups with domain-specialized safety models are a natural future direction.
\section{Conclusion}
\label{sec:conclusion}
\textbf{Synthesis: Toward Uncertainty-Aware Multi-Agent Moderation.}
Our empirical findings collectively establish that supervised LLM Performance Predictors enable a principled, cost-effective approach to human-AI collaboration in content moderation. The meta-model acts as an intelligent gating agent, coordinating between autonomous LLM agents and human reviewers by dynamically allocating tasks based on predicted error likelihood and operational constraints.

The LPP framework also opens research directions at the intersection of uncertainty and multi-agent coordination: (i) hierarchical escalation to reviewers with varying expertise, (ii) federated learning for privacy-preserving training across platforms, and (iii) integration with RLHF to couple uncertainty estimation with model improvement.

Ultimately, effective human-AI collaboration depends not only on accuracy but on systems that \emph{know what they don’t know}. By quantifying and attributing uncertainty, LPPs turn opaque LLM outputs into actionable signals, enabling reviewers to focus their expertise where it matters most.

\bibliographystyle{ACM-Reference-Format}
\bibliography{main}

\appendix
\section{Prompt Templates and Inference Configuration}
\label{app:prompts}

This appendix documents the schemas, prompt templates, and inference settings used to generate LLM outputs. Placeholders \verb+{{TEXT}}+, \verb+{{TRANSCRIPT}}+, \verb+{{THUMBNAIL}}+, \verb+{{VIDEO\FRAMES}}+, and \verb+{{CONCEPT_DEFINITION}}+ were filled dynamically for each query.

\subsection{Structured Output Schema}
\label{app:schema}

Models were instructed to emit a \texttt{JSON} object conforming to a predefined schema with a single integer outcome in: \texttt{0}~=~``no'', \texttt{1}~=~``yes'', \texttt{2}~=~``inconclusive\_evidence'', \texttt{3}~=~``inconclusive\_definition'', which are validated against a fixed schema; invalid (malformed) outputs trigger up to three deterministic retries.\\
The \texttt{ConceptClassification} object contains the final \texttt{outcome}, optional \verb+reasoning_steps+, self-reported confidence \verb+p_correct+, and a confidence \texttt{band}. \\
For reproducibility, the overall \verb+LLMClassification+ object and \verb+ReasoningStep+ are shown below.

\begingroup\small
\begin{verbatim}
class LLMClassification(BaseModel):
    classifications: Dict[ConceptEnum, ConceptClassification]

class ConceptClassification(BaseModel):
    outcome: Optional[OutcomeEnum]
    reasoning_steps: Optional[List[ReasoningStep]]
    p_correct: Optional[int]
    band: Optional[BandEnum]

class ReasoningStep(BaseModel):
    step_number: int
    description: str
\end{verbatim}
\endgroup

\myparagraph{Schema Clarifications}
All CoT prompts enforced exactly three \verb+reasoning_steps+. We used regular expressions to normalize the outcome tokens\\
\texttt{(yes|no|inconclusive\_(definition|evidence))} for consistent parsing. The \verb+p_correct+ field was snapped to the nearest multiple of 5 in \([0,100]\), and the \texttt{band} was validated against \verb+{VL, L, M, H, VH}+.

\subsection{Representative Prompt Templates}
\label{app:prompts-templates}

\myparagraph{Text-only Baseline (System Prompt)}
\begin{verbatim}
# ROLE AND GOAL
You are a meticulous Chief Marketing Officer (CMO) ...
...
\end{verbatim}

\myparagraph{Text-only User Prompt}
\begin{verbatim}
Please classify the following content:
--- CONTENT START ---
{{TEXT}}
--- CONTENT END ---
\end{verbatim}

\myparagraph{Multimodal User Prompt}
\begin{verbatim}
Analyze the following multimodal content.

--- START OF MULTIMODAL CONTENT ---
<VIDEO\FRAMES> {{VIDEO\FRAMES}} </VIDEO\FRAMES>
<THUMBNAIL> {{THUMBNAIL}} </THUMBNAIL>
<TRANSCRIPT> {{TRANSCRIPT}} </TRANSCRIPT>
<CONTENT_TEXT> {{TEXT}} </CONTENT_TEXT>
--- END OF MULTIMODAL CONTENT ---
\end{verbatim}
Depending on the base LLM, the visual input was provided either as a video URI or as a set of key frames; the placeholder \verb+{{VIDEO\FRAMES}}+ denotes this modality.
The CoT variants added a \texttt{REASONING FRAMEWORK} section to the system prompt to enforce a step-by-step thought process. The multimodal variants included placeholders for all four modalities and a protocol requiring a review of each.

\subsection{Inference Configuration}
\label{app:inference}

The following fixed decoding parameters were used for all API calls:
\begin{itemize}
    \item Decoding: \texttt{temperature=0}, \texttt{top-p=1}, \texttt{n=1}.
    \item Max output tokens: \texttt{8096}.
    \item Log-probabilities: top-20 per token.
\end{itemize}
Tokens were segmented into classification and reasoning spans based on the structured output. Assets (images, videos) were passed as Google Cloud Storage URIs, and malformed JSON responses (fewer than 2\% of total requests) were retried up to three times.

\begin{table*}[t]
\centering
\caption{\textbf{Catalog of LLM Performance Predictors (LPPs).} Features are grouped by access (gray-box vs. black-box). \textbf{A--B}: Outcome-level confidence metrics (top-$k$, schema-filtered). \textbf{C}: Log-probability and confidence margins. \textbf{D--E}: Reasoning-path features from CoT (perplexity, entropy). \textbf{F}: Self-reported confidence. \textbf{G}: Moderation-oriented abstention signals, distinguishing $\textit{aleatoric}$ (evidence) vs. $\textit{epistemic}$ (policy) uncertainty. Filtered variants restrict to schema labels $\mathcal{A}=\{0,1,2,3\}$; $\varepsilon=10^{-12}$ ensures stability.}
\small
\begin{tabular}{p{4.0cm} p{4.8cm} p{8.2cm}}
\toprule
\textbf{Feature Family} & \textbf{Mathematical Formula} & \textbf{Description / Intuition} \\
\midrule
\multicolumn{3}{l}{\textit{\textbf{A. Outcome Distribution: Unfiltered Top-$k$ (Gray-Box Access)}}} \\
\multicolumn{3}{p{17.0cm}}{\small \textbf{Context}: Given top-$k$ log-probabilities $\{\ell_1, \ldots, \ell_k\}$ for the outcome token (we use $k{=}5$), compute renormalized probabilities $\tilde{p}_i = \exp(\ell_i) / \sum_{j=1}^k \exp(\ell_j)$.} \\[0.5ex]
\hdashline
Entropy (base-2) & $H_2(\tilde{p}) = -\sum_{i=1}^k \tilde{p}_i \log_2 \tilde{p}_i$ & Uncertainty over the outcome distribution; higher values indicate greater indecision. \\
Normalized Entropy & $H_2^{\text{norm}}(\tilde{p}) = \frac{H_2(\tilde{p})}{\log_2 k}$ & Entropy scaled to $[0,1]$, invariant to $k$. \\
Effective Choices & $N_{\text{eff}} = 2^{H_2(\tilde{p})}$ & Number of equally likely outcomes consistent with the entropy (interpretable as perplexity). \\
Confidence Score (Top-$k$) & $C_{\text{top-}k} = 1 - H_2^{\text{norm}}(\tilde{p})$ & Complement of normalized entropy; ranges $[0,1]$ with 1 = fully confident. \\
Max Softmax Probability (MSP) & $\text{MSP} = \max_{i} \tilde{p}_i$ & Classic confidence baseline; lower values imply greater uncertainty. \\
Top-2 Margin & $\Delta p = \tilde{p}_{(1)} - \tilde{p}_{(2)}$ & Absolute separation between the most and second-most probable outcomes. \\
Top-2 Margin (Normalized) & $\Delta p^{\text{norm}} = \frac{\tilde{p}_{(1)} - \tilde{p}_{(2)}}{\max\{\tilde{p}_{(1)}, \varepsilon\}}$ & Relative margin; $\varepsilon{=}10^{-12}$ prevents division by zero. \\
Top-1/Top-2 Ratio & $R_{1/2} = \frac{\tilde{p}_{(1)}}{\max\{\tilde{p}_{(2)}, \varepsilon\}}$ & Confidence ratio between top two candidates. \\
\midrule
\multicolumn{3}{l}{\textit{\textbf{B. Outcome Distribution: Filtered to Valid Schema Labels (Gray-Box Access)}}} \\
\multicolumn{3}{p{17.0cm}}{\small \textbf{Context}: Request top-20 log-probabilities, filter to schema-valid labels $\mathcal{A}$ (binary: $\{0,1\}$; expanded: $\{0,1,2,3\}$), collapse token mass onto each label, renormalize to obtain $\tilde{p}^{(\mathcal{A})}$, then recompute metrics.} \\[0.5ex]
\hdashline
Filtered Entropy & $H_2(\tilde{p}^{(\mathcal{A})})$ & Uncertainty conditional on valid labels only; robust to spurious high-probability tokens. \\
Filtered Normalized Entropy & $H_2^{\text{norm}}(\tilde{p}^{(\mathcal{A})})$ & Filtered entropy scaled to $[0,1]$. \\
Filtered Effective Choices & $2^{H_2(\tilde{p}^{(\mathcal{A})})}$ & Effective number of schema-consistent outcomes. \\
Filtered Confidence Score & $1 - H_2^{\text{norm}}(\tilde{p}^{(\mathcal{A})})$ & Confidence over valid decision classes. \\
Filtered Top-2 Margin & $\Delta p^{(\mathcal{A})} = \tilde{p}^{(\mathcal{A})}_{(1)} - \tilde{p}^{(\mathcal{A})}_{(2)}$ & Margin between top two valid labels. \\
Filtered Top-2 Margin (Normalized) & $\frac{\tilde{p}^{(\mathcal{A})}_{(1)} - \tilde{p}^{(\mathcal{A})}_{(2)}}{\max\{\tilde{p}^{(\mathcal{A})}_{(1)}, \varepsilon\}}$ & Normalized filtered margin. \\
Filtered Top-1/Top-2 Ratio & $\frac{\tilde{p}^{(\mathcal{A})}_{(1)}}{\max\{\tilde{p}^{(\mathcal{A})}_{(2)}, \varepsilon\}}$ & Confidence ratio over valid labels. \\
\midrule
\multicolumn{3}{l}{\textit{\textbf{C. Log-Probability Margin (Gray-Box Access)}}} \\
\hdashline
Log-Odds Margin (Top-2) & $\Delta_\ell = \ell_{(2)} - \ell_{(1)}$ & Difference in log-space; more negative = stronger preference for top outcome. \\
Normalized Log-Odds Margin & $\Delta_\ell^{\text{norm}} = \frac{\ell_{(2)} - \ell_{(1)}}{\ell_{(2)}}$ & Relative Margin; stabilizes comparison across models/prompts.\\
Filtered Log-Odds Margin & $\Delta_\ell^{(\mathcal{A})} = \ell^{(\mathcal{A})}_{(2)} - \ell^{(\mathcal{A})}_{(1)}$ & Margin computed over filtered valid labels. \\
Filtered Normalized Top 2 Margin & $\frac{\ell^{(\mathcal{A})}_{(2)} - \ell^{(\mathcal{A})}_{(1)}}{\ell^{(\mathcal{A})}_{(2)}}$ & Normalized version for filtered labels. \\
\midrule
\multicolumn{3}{l}{\textit{\textbf{D. Reasoning Sequence-Level Features (Gray-Box Access, CoT Required; not used in final reported results})}} \\
\multicolumn{3}{p{17.0cm}}{\small \textbf{Context}: Given a CoT reasoning sequence of $T$ tokens $\{y_1, \ldots, y_T\}$ with per-token log-probabilities $\{\log p(y_t)\}_{t=1}^T$.} \\[0.5ex]
\hdashline
Sequence Negative Log-Likelihood & $\text{NLL} = -\sum_{t=1}^{T} \log p(y_t)$ & Total surprise of the reasoning sequence; higher = less confident generation. \\
Perplexity (Natural Base) & $\text{PPL} = \exp\!\left(-\frac{1}{T} \sum_{t=1}^{T} \log p(y_t)\right)$ & Standard perplexity metric; higher values indicate less fluent reasoning. \\
\midrule
\multicolumn{3}{l}{\textit{\textbf{E. Reasoning Token-Level Distributional Features (Gray-Box Access, CoT Required; not used in final reported results})}} \\
\multicolumn{3}{p{17.0cm}}{\small \textbf{Context}: For each reasoning token $y_t$, compute per-token entropy $h_t = -\sum_j \tilde{q}_{tj} \log_2 \tilde{q}_{tj}$ where $\tilde{q}_{tj}$ are renormalized top-$k$ probabilities. Aggregate over $T$ tokens.} \\[0.5ex]
\hdashline
Mean Token Entropy & $\bar{h} = \frac{1}{T} \sum_{t=1}^T h_t$ & Average per-token uncertainty during generation. \\
Token Entropy Quantiles & $Q_q(\{h_t\}),\ q \in \{0, 0.25, 0.5, 0.75, 1.0\}$ & Distributional shape of per-token uncertainties . \\
Token Probability Quantiles & $Q_q(\{p(y_t)\}),\ q \in \{0, 0.25, 0.5, 0.75, 1.0\}$ & Distributional quantiles of per-token likelihoods. \\
\midrule
\multicolumn{3}{l}{\textit{\textbf{F. Verbalized (Self-Reported) Confidence (Black-Box Compatible)}}} \\
\hdashline
Reported Confidence (Scalar) & $\hat{c} \in [0, 100]$ (normalized to $[0,1]$) & LLM's explicit self-assessment of confidence, extracted from structured output. \\
Confidence Bands (One-Hot) & $\mathbb{I}\{\text{VL, L, M, H, VH}\}$ & Coarse-grained verbalization: Very Low, Low, Medium, High, Very High.\\
\midrule
\multicolumn{3}{l}{\textit{\textbf{G. Uncertainty Attribution Indicators (Black-Box Compatible)}}} \\
\hdashline
Evidence-Deficit Indicator (Binary) &
$\mathbb{I}\{\text{outcome} = 2\}$ &
1 if the LLM abstained due to insufficient evidence (aleatoric); 0 otherwise. \\
Policy-Gap Indicator (Binary) &
$\mathbb{I}\{\text{outcome} = 3\}$ &
1 if the LLM abstained due to a definition/policy gap (epistemic); 0 otherwise. \\
\bottomrule
\end{tabular}
\label{app:full_lpps}
\end{table*}
\subsection{Cost-Ratio Sensitivity}
\label{app:cost_ratio_sensitivity}

We recompute the relative expected cost at the fixed operating point $\tau^\star$ under cost-ratio variation
$r=c_{\text{rev}}/c_{\text{mis}}\in[0.4,0.9]$ using the confusion-derived counts $(FP,TN,FN)$ and the cost
definition in \S3.3: $C/c_{\text{mis}} = FP + (r-1)\cdot TN + r\cdot FN$.
Table~\ref{tab:cost_ratio_sensitivity} reports values at representative ratios $\{0.4,0.64,0.9\}$ spanning the swept range.

\begin{table*}[t]
\centering
\caption{Cost-ratio sensitivity analysis. Relative expected cost $C(\tau^\star)/c_{\text{mis}}$ recomputed for cost ratios $r = c_{\text{rev}}/c_{\text{mis}} \in [0.4, 0.9]$, reported at representative values $\{0.4, 0.64, 0.9\}$ (range endpoints and baseline).}
\label{tab:cost_ratio_sensitivity}

\small
\setlength{\tabcolsep}{4pt}
\renewcommand{\arraystretch}{1.05}

\begin{minipage}[t]{0.485\textwidth}
\centering
\subcaption{OpenAI Moderation Dataset}
\begin{tabularx}{\linewidth}{@{}>{\raggedright\arraybackslash}Xccc@{}}
\toprule
LLM & $r=0.4$ & $r=0.64$ & $r=0.9$ \\
\midrule
gpt-4o-mini            &  78 & 38 & 178 \\
gpt-4o                 & 106 & 51 & 190 \\
gpt-4.1-mini           & 163 & 45 & 244 \\
gemini-2.5-flash-lite  & 187 & 77 & 285 \\
gemini-2.0-flash-lite  &  97 & 41 & 194 \\
gemini-2.0-flash-001   & 201 & 69 & 289 \\
QWEN3-14B              & 121 & 56 & 212 \\
QWEN3-32B              & NA  & NA & NA  \\
LLAMA32-11B            & NA  & NA & NA  \\
\bottomrule
\end{tabularx}
\end{minipage}
\hfill
\begin{minipage}[t]{0.485\textwidth}
\centering
\subcaption{Multimodal Moderation Dataset}
\begin{tabularx}{\linewidth}{@{}>{\raggedright\arraybackslash}Xccc@{}}
\toprule
LLM & $r=0.4$ & $r=0.64$ & $r=0.9$ \\
\midrule
gpt-4o-mini            & 72 & 22 &  96 \\
gpt-4o                 & 18 & 29 &  63 \\
gpt-4.1-mini           & 20 & 30 &  65 \\
gemini-2.5-flash-lite  &  0 & 40 &  54 \\
gemini-2.0-flash-lite  & 54 & 45 & 100 \\
gemini-2.0-flash-001   &  0 & 34 &  44 \\
QWEN3-14B              & 53 & 36 &  93 \\
QWEN3-32B              & 31 & 47 &  85 \\
LLAMA32-11B            & 71 & 39 & 108 \\
\bottomrule
\end{tabularx}
\end{minipage}

\end{table*}
\end{document}